%% file: main.tex
\documentclass{article}

\PassOptionsToPackage{numbers, compress}{natbib}

\usepackage[final]{nips_2017}

\usepackage[utf8]{inputenc} 
\usepackage[T1]{fontenc}    
\usepackage[breaklinks=true,bookmarks=false,colorlinks]{hyperref}
\usepackage{url}            
\usepackage{booktabs}       
\usepackage{amsfonts}       
\usepackage{nicefrac}       
\usepackage{microtype}      

\usepackage[pdftex]{graphicx}
\usepackage{multirow}
\usepackage{authblk}
\usepackage{hhline}
\newcommand{\netshort}{SINet-Caption}
\newcommand{\Myhat}[1]{\expandafter\hat#1}

\title{\Large{Grounded Objects and Interactions for Video Captioning}}

\makeatletter
\renewcommand\AB@affilsepx{, \protect\Affilfont}
\makeatother

\author[1]{Chih-Yao Ma}
\author[2]{Asim Kadav}
\author[2]{Iain Melvin}
\author[3]{Zsolt Kira}
\author[1]{Ghassan AlRegib}
\author[2]{Hans Peter Graf}
\affil[1]{\small Georgia Institute of Technology}
\affil[2]{\small NEC Laboratories America}
\affil[3]{\small Georgia Tech Research Institute}

\begin{document}

\maketitle

\input{sections/abstract}
\input{sections/intro}
\input{sections/model}
\input{sections/eval}
\input{sections/supplementary}

{\small
\bibliographystyle{abbrvnat}
\bibliography{egbib}
}

\end{document}

%% file: sections/abstract.tex
\begin{abstract}
We address the problem of video captioning by grounding language generation on object interactions in the video.
Existing work mostly focuses on overall scene understanding with often limited or no emphasis on object interactions to address the problem of video understanding.
In this paper, we propose \netshort\ that learns to generate captions grounded over higher-order interactions between arbitrary groups of objects for fine-grained video understanding. 
We discuss the challenges and benefits of such an approach.
We further demonstrate state-of-the-art results on the ActivityNet Captions dataset using our model, \netshort\ based on this approach.
\end{abstract}

%% file: sections/intro.tex
\section{Introduction}

\begin{figure}[t]
    \centering
    \includegraphics[width=1\textwidth]{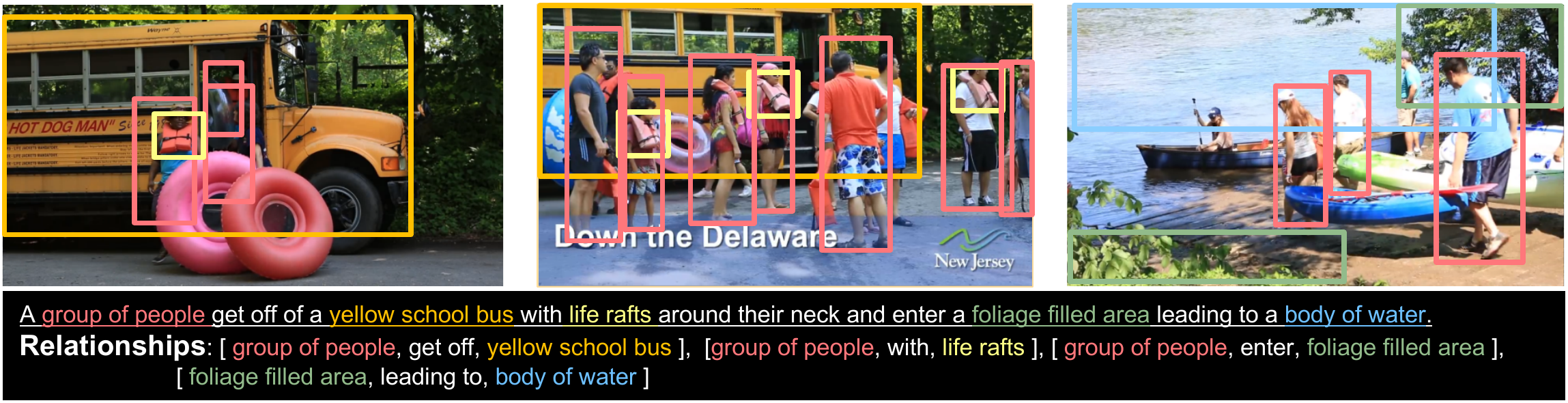}
    \caption{
    Video captions are composed of multiple visual relationships and interactions. 
    We detect higher-order object interactions and use them as basis for video captioning. 
    }
    \label{fig:concept}
    \vspace{-0.1in}
\end{figure}

Video understanding using natural language processing to generate captions for videos has been regarded as a crucial step towards machine intelligence. 
However, like other video understanding tasks addressed using deep learning, initial work on video captioning has focused on learning compact representations combined over time that is used as an input to a decoder, e.g. LSTM, at the beginning or at each word generation stage to generate a caption for the target video~\cite{pan2016jointly, venugopalan2015sequence, venugopalan2014translating}. 
This has been improved by using spatial and temporal attention mechanisms ~\cite{song2017hierarchical, yao2015describing, yu2016video} and/or semantics attribute methods~\cite{gan2016semantic, pan2016video, shen2017weakly, yu2017end}. 
These methods do not ground their predictions on object relationships and interactions, i.e. they largely focus on overall scene understanding or certain aspects of the video at best.
However, modeling visual relationships and object interactions in a scene is a crucial form of video understanding as shown in Figure~\ref{fig:concept}.

There has been considerable work that detect visual relationships in images using separate branches in a ConvNet to explicitly model object, human, and their interactions~\cite{chao2017learning, gkioxari2017detecting}, using scene graphs~\cite{johnson2015image, li2017scene, liang2017deep, xu2017scene} and by pairing different objects in the scene~\cite{dai2017detecting, hu2016modeling, santoro2017simple, zhang2017visual}. While these models can successfully detect visual relationships for images, these methods are intractable
in the video domain making it inefficient if not impossible to detect all relationships across all individual object pairs~\cite{zhang2017relationship}. As a result, past work has at best focused on pairwise relationships on limited datasets~\cite{lea2016segmental, ni2014multiple}.

In this paper, we first hypothesize using these interactions as basis for caption generation. We then describe our method to achieve this
goal by using an Region Proposal Network (RPN) to extract ROIs from videos and learning their interactions efficiently using dot product attention. 
Our model, \textbf{\netshort} efficiently explores and grounds caption generation over interactions between arbitrary subgroups of objects, the members of which are determined by a learned attention mechanism. In our results,
we demonstrate how to exploit both overall image context (coarse-grained) and higher-order object interactions (fine-grained) in the spatiotemporal domain for video captioning, as illustrated in Figure~\ref{fig:coarse-fine}. We obtain state-of-the-art results on video
captioning over the challenging ActivityNet Captions dataset.

%% file: sections/model.tex
\begin{figure}[b]
    \centering
    \includegraphics[width=0.75\textwidth]{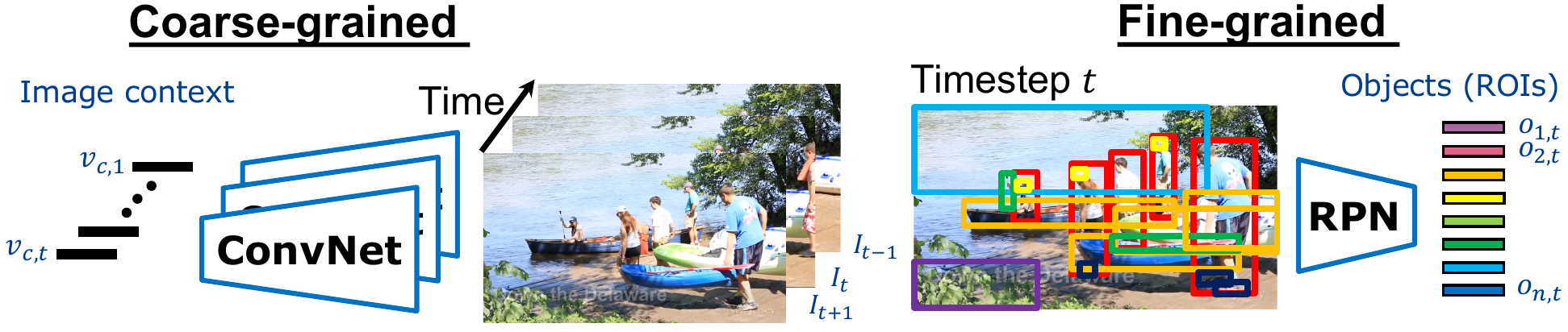}
    \caption{
    We exploit both coarse- (overall image) and fine-grained (object) visual representations for each video frame. 
    }
    \label{fig:coarse-fine}
\end{figure}

\section{Video captioning model}
The proposed \textbf{\netshort} first attentively models object inter-relationships and discovers the higher-order interactions for a video. 
The detected higher-order object interactions (fine-grained) and overall image representation (coarse-grained) are then temporally attended as visual cue for each word generation.

\subsection{Higher-order object interactions}
\label{sec:obj-interaction}

\textbf{Problem statement:}
We define \textit{objects} to be a certain region in the scene that might be used to determine the visual relationships and interactions. 
Each object representation can be directly obtained from an RPN and further encoded into an object feature, as shown in Figure~\ref{fig:coarse-fine}. 
Note that we do not encode class information from the object detector since there exists cross-domain problem, and we may miss some objects that are not detected by pre-trained object detector. 
Also, we do not know the corresponding object across time since linking objects through time may not be suitable if the video sequence is long and computationally expensive.  
Our objective is to efficiently detect higher-order interactions---\textit{interactions beyond pairwise objects}---from these rich yet unordered object representations reside in a high-dimensional space that spans across time. 

In the simplest setting, an interaction between objects in the scene can be represented as combining individual object information. 
For example, one method is to add the learnable representations and project these representations into a high-dimensional space where the object interactions can be exploited by simply summing up the object representations~\cite{santoro2017simple}. 
Another approach which has been widely used with images is pairing all possible object candidates (or subject-object pairs)~\cite{chao2017learning, dai2017detecting,hu2016modeling,zhang2017visual}. 
However, this is infeasible for video, since a video typically contains hundreds of frame and the set of object-object pairs is too large to fully represent. 
Detecting object relationships frame by frame is computationally expensive, and the temporal reasoning of object interactions is not used.

\begin{figure}[t]
    \centering
    \includegraphics[width=1\textwidth]{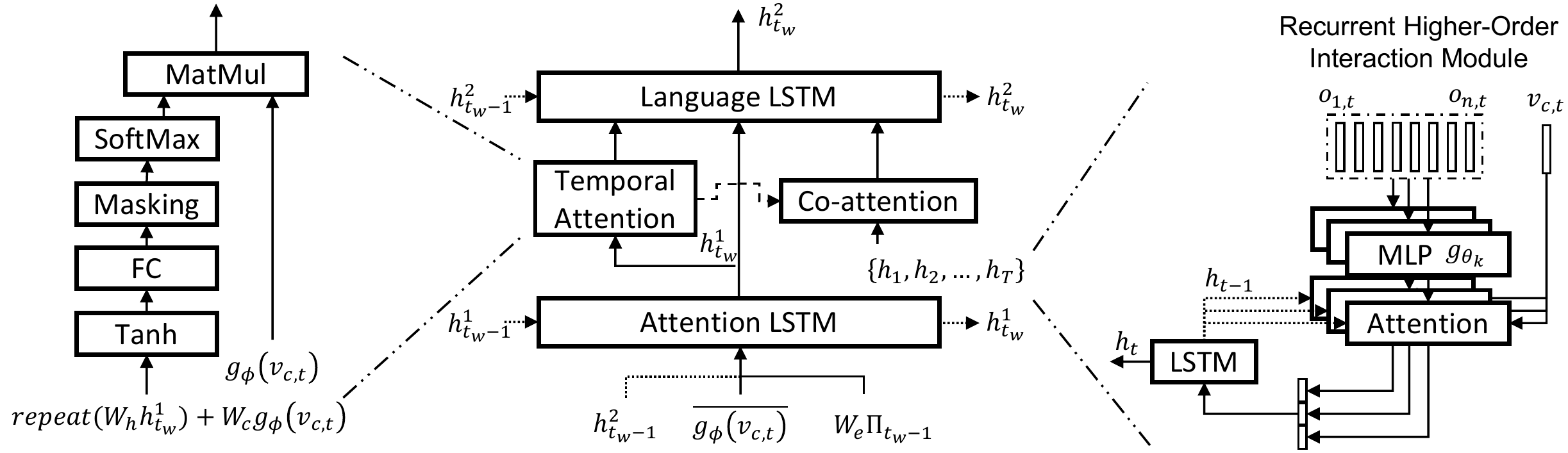}
    \caption{
    Overview of the proposed \netshort\ for video captioning. 
    }
    \label{fig:model-caption}
    \vspace{-0.1in}
\end{figure}

\textbf{Recurrent Higher-Order Interaction Module:}
To overcome these issues, we propose a generic recurrent module for detecting higher-order object interactions for fine-grained video understanding problem, as illustrated in Figure~\ref{fig:model-caption} (right). 
The proposed recurrent module dynamically selects object candidates which are important to describe video content. The combinations of these selected objects are then concatenated to model higher-order interaction using group to group or triplet groups of objects.


Formally, the current image representation $v_{c,t}$ and previous object interaction representation $h_{t-1}$ are first projected to introduce learnable weights. The projected $v_{c,t}$ and $h_{t-1}$ are then repeated and expanded $n$ times (the number of objects at time $t$). We directly combine this information with projected objects via matrix addition and use it as input to dot-product attention.
In short, the attention is computed using inputs from current (projected) object features, overall image visual representation, and previously discovered object interactions, which provide the attention mechanism with maximum context. 
Specifically, the input to the attention can be define as: $X_{k} = repeat(W_{h_k} h_{t-1} + W_{c_k} v_{c,t}) + g_{\theta_k} (O_t)$, and the attention weights over all objects can be define as: $\alpha_{k} = softmax(\frac{{X_{k}}^\top X_{k}}{\sqrt{d_\theta}})$
, where $W_{h_k}$ and $W_{c_k}$ 
are learned weights for $h_{t-1}$ and $v_{c,t}$. 
$g_{\theta_k}$ is a Multi-Layer Perceptron (MLP) with parameter $\theta_k$,
$d_{\theta}$ is the dimension of last fully-connected layer of $g_{\theta_k}$,
$X_{k}$ is the input to $k$th attention module,
and $\sqrt{d_\theta}$ is a scaling factor.
We omit the bias term for simplicity. 

The attended object feature at time $t$ is then calculated as mean-pooling on weighted objects:
$v_{{o, t}}^k = \overline{\alpha_k\ (g_{\theta_k} (O_t))^\top}$.
The output $v_{{o, t}}^k$ is a single feature vector representation which encodes the $k$th object inter-relationships of a video frame at time $t$, and $k$ ranges from $\big\{1, 2, ..., K\big\}$ representing the number of groups for inter-relationships. 

Finally, the selected object candidates $v_{{o, t}}^k$ are then concatenated and used as the input to a LSTM cell: 
$h_{t} = LSTM (v_{{o, t}}^1 \Vert v_{{o, t}}^2 \Vert ... \Vert v_{{o, t}}^K)$.
The output $h_{t}$ is then defined as the higher-order object interaction representation at current time $t$.
The sequence of hidden state of the LSTM cell $\mathbf{h} = \big(h_1, h_2, ..., h_T\big)$ are the representations of higher-order object interactions for each timestep.

\subsection{Video captioning with coarse- and fine-grained}
We now describe how our \textbf{\netshort}\ exploit both coarse-grained (overall image representation) and fine-grained (higher-order object interactions) for video captioning. 

The \netshort\ is inspired by prior work using hierarchical LSTM for captioning task~\cite{anderson2017bottom,song2017hierarchical}, and we extend it with the proposed recurrent higher-order interaction module so that the model can leverage the detected higher-order object interactions, as shown in Figure~\ref{fig:model-caption}. 
The model consist of two LSTM layers: Attention LSTM and Language LSTM.

\textbf{Attention LSTM:}
The Attention LSTM fuses the previous hidden state output of Language LSTM $h_{t_w-1}^2$, overall representation of the video, and the input word at time $t_w-1$ to generate the hidden representation for the following attention module.
Formally, the input to Attention LSTM can be defined as: $x_{t_w}^1 = h_{t_w-1}^2 \ \Vert \ \overline{g_{\phi} (v_{c,t}}) \ \Vert \ W_e \Pi_{t_w-1}$,
where $\overline{g_{\phi}(v_{c,t})}$ is the projected and mean-pooled image features,
$g_\phi$ is a MLP with parameter $\phi$,
$W_e \in \mathbb{R}^{E \times \Sigma}$ is a word embedding matrix for a vocabulary of size $\Sigma$,
and $\Pi_{t_w-1}$ is one-hot encoding of the input word at time $t_w-1$.
Note that $t$ is the video time, and $t_w$ is the timestep for caption generation.

\textbf{Temporal attention:}
The input for computing the temporal attention is the combination of output of the Attention LSTM $h_{t_w}^1$ and projected image features $g_{\phi} (v_{c,t})$, and the attention is computed using a simple \textit{tanh} function and a fully-connected layer to attend to projected image features $g_{\phi} (v_{c,t})$, as illustrated in Figure~\ref{fig:model-caption}.
Specifically, the input can be defined as $x_a = repeat(W_h h_{t_w}^1) + W_c g_{\phi} (v_{c,t})$, and the temporal attention can be obtained by $\alpha_{temp} = softmax({w_a}^\top tanh(x_a))$
, where $W_h \in \mathbb{R}^{d_{g_{\phi}} \times d_h}$ and $W_c \in \mathbb{R}^{d_{g_{\phi}} \times d_{g_{\phi}}}$ are learned weights for $h_{t}^1$ and $g_{\phi} (v_{c,t})$. 
$d_{g_{\phi}}$ is the dimension of last fully-connected layer of $g_{\phi}$.

\textbf{Co-attention:}
We directly apply the temporal attention obtained from image features on object interaction representations $\mathbf{h} = \big( h_1, h_2, ..., h_T\big)$ (see Sec~\ref{sec:obj-interaction} for details).

\textbf{Language LSTM:}
Finally, the Language LSTM takes the concatenation of output of the Attention LSTM $h_{t_w}^1$, attended video representation $\Myhat{v_{c,t_w}}$, and co-attended object interactions $\Myhat{h_{t_w}}$ as input: $x_{t_w}^2 = h_{t_w}^1 \ \| \ \Myhat{v_{c,t_w}} \ \Vert \ \Myhat{h_{t_w}}$. 
The output of Language LSTM is then used to generate words via fully-connected and softmax layer.

%% file: sections/eval.tex
\begin{table}[t]
    \centering
    \caption{
        METEOR~\cite{banerjee2005meteor}, ROUGE-L~\cite{lin2004rouge}, CIDEr-D~\cite{vedantam2015cider}, and BLEU@N~\cite{papineni2002bleu} scores on the ActivityNet Captions test and validation set. All methods use ground truth proposal except LSTM-A$_3$~\cite{Yao2017LSTMA3}.
        Our results with ResNeXt spatial features use videos sampled at maximum 1 FPS only.
        }
    \label{table:caption-eval}
    \small
    \begin{tabular}{lccccccc}
    Method  & B@1   & B@2  & B@3 & B@4    & R    & M    & C   \\ \hline
    \multicolumn{1}{c}{\textbf{Test set}} & & & & & & & \\ \hline
    LSTM-YT~\cite{venugopalan2014translating} (C3D)  & 18.22    & 7.43  & 3.24  & 1.24      & -     & 6.56   &14.86 \\
    S2VT~\cite{venugopalan2015sequence} (C3D)  & 20.35    & 8.99  & 4.60  & 2.62      & -     & 7.85   & 20.97 \\
    H-RNN~\cite{yu2016video} (C3D)  & 19.46    & 8.78  & 4.34  & 2.53      & -     & 8.02   & 20.18 \\
    S2VT + full context~\cite{krishna2017dense} (C3D)  & 26.45    & 13.48  & 7.21  & 3.98      & -     & 9.46   & 24.56 \\
    \multirow{2}{*}{\begin{tabular}[c]{@{}c@{}}LSTM-A$_3$ + policy gradient + retrieval~\cite{Yao2017LSTMA3}\\ (ResNet + P3D ResNet~\cite{Qiu_2017_ICCV})\end{tabular}} & \multirow{2}{*}{-} & \multirow{2}{*}{-} & \multirow{2}{*}{-} & \multirow{2}{*}{-} & \multirow{2}{*}{-} & \multirow{2}{*}{12.84} & \multirow{2}{*}{-} \\
    &   &   &   &   &   &   &   \\ \hline 
    \multicolumn{1}{c}{\textbf{Validation set (Avg. 1st and 2nd)}} & & & & & & & \\ \hline
    LSTM-A$_3$ (ResNet + P3D ResNet)~\cite{Yao2017LSTMA3}   & - & - & -     & \textbf{3.38}  & 13.27 & 7.71  & 16.08 \\
    \multirow{2}{*}{\begin{tabular}[c]{@{}c@{}}LSTM-A$_3$ + policy gradient + retrieval~\cite{Yao2017LSTMA3}\\ (ResNet + P3D ResNet~\cite{Qiu_2017_ICCV})\end{tabular}} & \multirow{2}{*}{-} & \multirow{2}{*}{-} & \multirow{2}{*}{-} & \multirow{2}{*}{3.13} & \multirow{2}{*}{14.29} & \multirow{2}{*}{8.73} & \multirow{2}{*}{14.75} \\
    &   &   &   &   &   &   &   \\ 
    \netshort\ | img (C3D)                         & 17.18 & 7.99  & 3.53  & 1.47  & 18.78 & 8.44  & 38.22 \\
    \netshort\ | img (ResNeXt)                     & 18.81 & 9.31  & 4.27  & 1.84  & 20.46 & 9.56  & 43.12 \\
    \netshort\ | obj (ResNeXt)                     & 19.07 & 9.48  & 4.38  & 1.92  & 20.67 & 9.56  & 44.02 \\
    \netshort\ | img+obj | no co-attn (ResNeXt)      & \textbf{19.93} & 9.82  & \textbf{4.52}  & 2.03  & 21.08 & 9.79  & 44.81 \\
    \netshort\ | img+obj (ResNeXt)     & 19.78    & \textbf{9.89}   & \textbf{4.52}   & 1.98 & \textbf{21.25}    & \textbf{9.84} & \textbf{44.84} 
\end{tabular}
\vspace{-0.1in}
\end{table}

\section{Evaluation on ActivityNet Captions}
We use the ActivityNet Captions dataset for evaluating \netshort. 
The ground truth temporal proposals are used to segment videos, i.e. we treat each video segment independently since our focus in this work is modeling object interactions for video captioning rather than on temporal proposals (please refer to supplementary material for details on dataset and implementation). 
All methods in Table~\ref{table:caption-eval} use ground truth temporal proposal, except LSTM-A$_3$~\cite{Yao2017LSTMA3}. 

For a fair comparison with prior methods that use C3D features, we report results with both C3D and ResNeXt spatial features.
Since there is no prior result reported on the validation set, we compare our method via LSTM-A$_3$~\cite{Yao2017LSTMA3} which reports results on the validation and test sets. 
This allows us to indirectly compare with methods reported on the test set. 
As shown in Table~\ref{table:caption-eval}, while LSTM-A$_3$ clearly outperforms other methods on the test set with a large margin, our method shows better results on the validation sets across nearly all language metrics. 
Note that we do not claim our method to be superior to LSTM-A$_3$ because of two fundamental differences. 
First, they do not rely on ground truth temporal proposals. 
Second, they use features extracted from a ResNet fine-tuned on Kinetics and another P3D ResNet~\cite{Qiu_2017_ICCV} fine-tuned on Sports-1M, whereas we only use an ResNeXt-101 fine-tuned on Kinetics sampled at maximum 1 FPS. 
Utilizing more powerful feature representations can improve the prediction accuracy by a large margin on video captioning tasks. This also corresponds to our experiments with C3D and ResNeXt features, where the proposed method with ResNeXt features performs significantly better than C3D features.
We observed that using only the detected object interactions shows slightly better performance than using only overall image representation. 
This demonstrates that even though the \netshort\ is not aware of the overall scene representation, it achieves similar performance relying on only the detected object interactions. 
By combining both, the performance further improved across all evaluation metrics, with or without co-attention. 

In conclusion, we introduce a generic recurrent module to detect higher-order object interactions for video understanding task. We demonstrate on ActivityNet Captions that the proposed \netshort\ exploiting both higher-order object interactions (fine-grained) and overall image representation (coarse-grained) outperforms prior work by a substantial margin.

%% file: sections/supplementary.tex
\section*{Supplementary Materials}

\subsection*{Dataset and Implementation} 
\textbf{ActivityNet Captions dataset:}
We use the new ActivityNet Captions for evaluating \netshort.  
The ActivityNet Captions dataset contains 20k videos and has total 849 video hours with 100k total descriptions. 
We focus on providing the fine-grained understanding of the video to describe video events with natural language, as opposed to identifying the temporal proposals. 
We thus use the ground truth temporal segments and treat each temporal segment independently. 
We use this dataset over other captioning datasets because ActivityNet Captions is action-centric, as opposed to object-centric~\cite{krishna2017dense}. 
This fits our goal of detecting higher-order object interactions for understanding human actions.
Following the same procedure as in \cite{krishna2017dense}, all sentences are capped to be a maximum length of 30 words. 
We sample predictions using beam search of size 5 for captioning. 
While the previous work sample C3D features every 8 frames~\cite{krishna2017dense}, we only sampled video at maximum 1 FPS. Video segments longer than 30 seconds are evenly sampled at maximum 30 samples. 


\textbf{Image feature:}
We pre-trained a ResNeXt-101 on the Kinetics dataset~\cite{kay2017kinetics} sampled at 1 FPS (approximately 2.5 million images), and use it to extract image features.
We use SGD with Nesterov momentum as the optimizer. 
The initial learning rate is $1e-4$ and automatically drops by 10x when validation loss saturated for 5 epochs. 
The weight decay is $1e-4$ and the momentum is 0.9, and the batch size is 128. 
We follow the standard data augmentation procedure by randomly cropping and horizontally flipping the video frames during training. 
When extracting image features, the smaller edge of the image is scaled to 256 pixels and we crop the center of the image as input to the fine-tuned ResNeXt-101.
Each image feature is encoded to a 2048-d feature vector. 

\textbf{Object feature:}
We generate the object features by first obtaining the coordinates of ROIs from a Deformable R-FCN~\cite{dai2017deformable} with ResNet-101~\cite{he2015deep} as the backbone architecture. 
The Deformable R-FCN was trained on MS-COCO train and validation dataset~\cite{lin2014microsoft}. 
We set the IoU threshold for NMS to be 0.2. 
For each of the ROIs, we extract features using ROI coordinates and adaptive max-pooling from the same ResNeXt-101 model pre-trained on Kinetics. 
The resulting object feature for each ROI is a 2048-d feature vector. 
ROIs are ranked according to their ROI scores. We select top 15 objects only. 
We do not use object class information since there is a cross-domain problem and we may miss some objects that were not detected.
For the same reason, the bounding-box regression process is not performed here since we do not have the ground-truth bounding boxes.

\textbf{Training:}
We train the proposed \netshort\ with ADAM optimizer. 
The initial learning rate is set to $1e-3$ and automatically drops by 10x when validation loss is saturated. 
The batch sizes is $32$. 

\subsection*{ActivityNet Captions on 1st and 2nd validation set}
We report the performance of \netshort\ on both 1st and 2nd validation set in Table~\ref{table:caption-two-eval}.
\begin{table*}[ht]
    \centering
    \caption{
        METEOR, ROUGE-L, CIDEr-D, and BLEU@N scores on the ActivityNet Captions 1st and 2nd validation set. All methods use ground truth temporal proposal, and out results are evaluated using the code provided in \cite{krishna2017dense} with $tIoU=0.9$.
        Our results with ResNeXt spatial features use videos sampled at maximum 1 FPS only.
        }
    \label{table:caption-two-eval}
    \small
    \begin{tabular}{lccccccc}
    Method  & B@1   & B@2  & B@3 & B@4    & R    & M    & C   \\ \hline
    \multicolumn{1}{c}{\textbf{1st Validation set}} & & & & & & & \\ \hline
    \netshort\ | img (C3D)    & 16.93 & 7.91 & 3.53  & 1.58  & 18.81 & 8.46  & 36.37 \\
    \netshort\ | img (ResNeXt)    & 18.71 & 9.21  & 4.25  & 2.00  & 20.42 & 9.55  & 41.18 \\
    \netshort\ | obj (ResNeXt)    & 19.00 & 9.42  & 4.29  & 2.03  & 20.61 & 9.50  & 42.20 \\
    \netshort\ | img+obj | no co-attn (ResNeXt)    & \textbf{19.89} & 9.76  & 4.48  & 2.15  & 21.00 & 9.62  & 43.24 \\
    \netshort\ | img+obj (ResNeXt)     & 19.63    & \textbf{9.87}   & \textbf{4.52}   & \textbf{2.17} & \textbf{21.22}    & \textbf{9.73} & \textbf{44.14} \\ \hline
    \multicolumn{1}{c}{\textbf{2nd Validation set}} & & & & & & & \\ \hline
    \netshort\ | img (C3D)    & 17.42 & 8.07 & 3.53  & 1.35  & 18.75 & 8.41  & 40.06 \\
    \netshort\ | img (ResNeXt)    & 18.91 & 9.41  & 4.28  & 1.68  & 20.49 & 9.56  & 45.05 \\
    \netshort\ | obj (ResNeXt)    & 19.14 & 9.53  & 4.47  & 1.81  & 20.73 & 9.61  & 45.84 \\
    \netshort\ | img+obj | no co-attn (ResNeXt)    & \textbf{19.97} & 9.88  & \textbf{4.55}  & \textbf{1.90}  & 21.15 & \textbf{9.96}  & \textbf{46.37} \\
    \netshort\ | img+obj (ResNeXt)     & 19.92    & \textbf{9.90}   & 4.52   & 1.79 & \textbf{21.28}    & 9.95 & 45.54
\end{tabular}
\end{table*}

\subsection*{Qualitative Analysis}
\begin{figure}[t]
    \centering
    \includegraphics[width=1\textwidth]{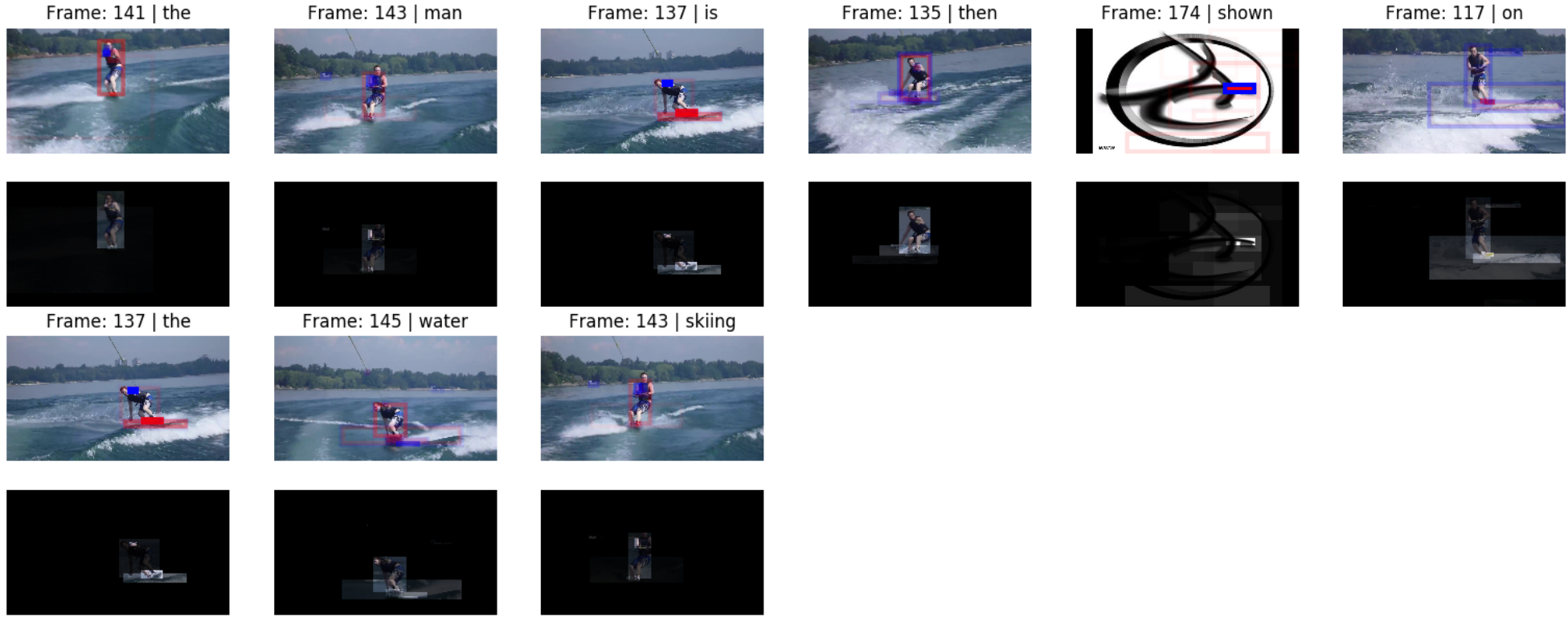}
    \caption{
    \textit{The man is then shown on the water skiing.}
    We can see that the proposed \netshort\ often focus on the person and the wakeboard, and most importantly it highlight the interaction between the two, i.e. the person steps on the wakeboard. 
    }
    \label{fig:water_surfing}
    \vspace{-0.1in}
\end{figure}

\begin{figure}[t]
    \centering
    \includegraphics[width=1\textwidth]{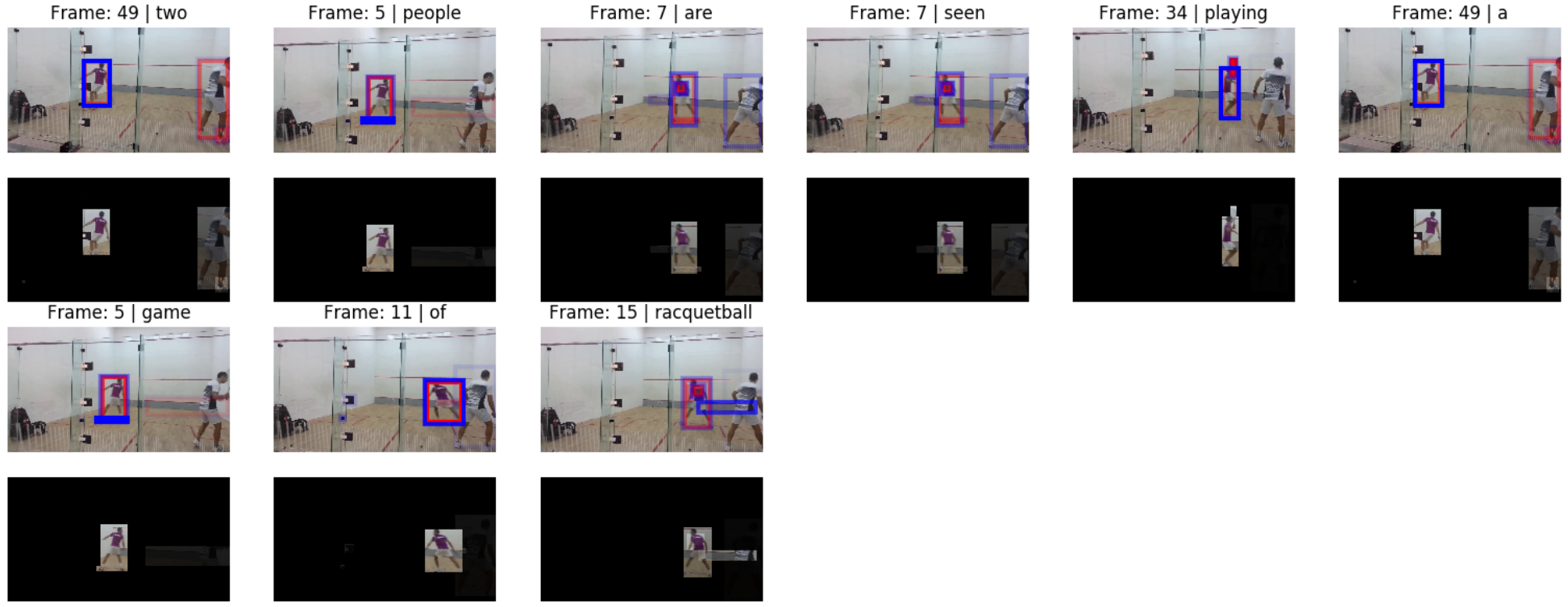}
    \caption{
    \textit{Two people are seen playing a game of racquetball.}
    The \netshort\ is able to identify that two persons are playing the racquetball and highlight the corresponding ROIs in the scene. 
    }
    \label{fig:racquetball}
    \vspace{-0.1in}
\end{figure}

\begin{figure}[t]
    \centering
    \includegraphics[width=1\textwidth]{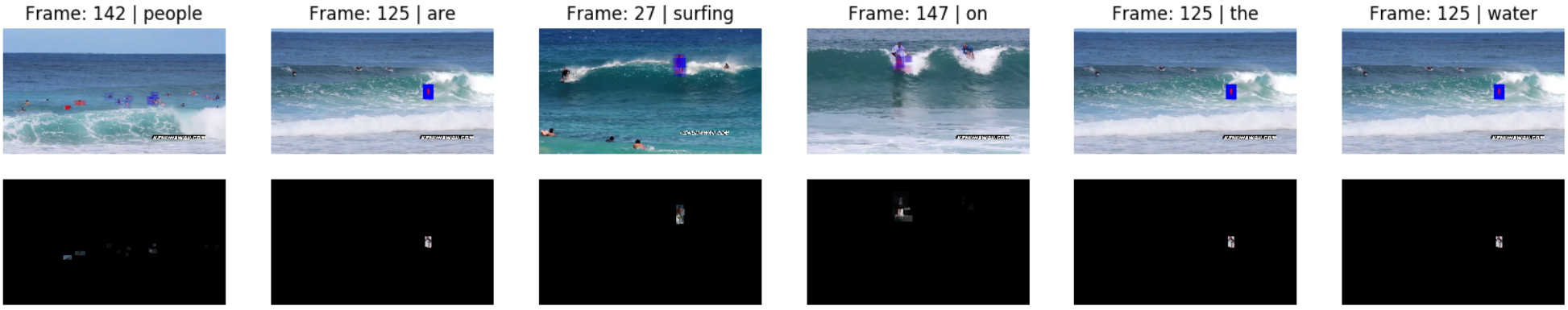}
    \caption{
    \textit{People are surfing on the water.}
    At the beginning, the \netshort\ identify multiple people are in the ocean. The person who is surfing on the water is then successfully identified, and the rest of irrelevant objects and background are completely ignored.
    }
    \label{fig:water_surfing_1}
    \vspace{-0.1in}
\end{figure}

\begin{figure}[t]
    \centering
    \includegraphics[width=1\textwidth]{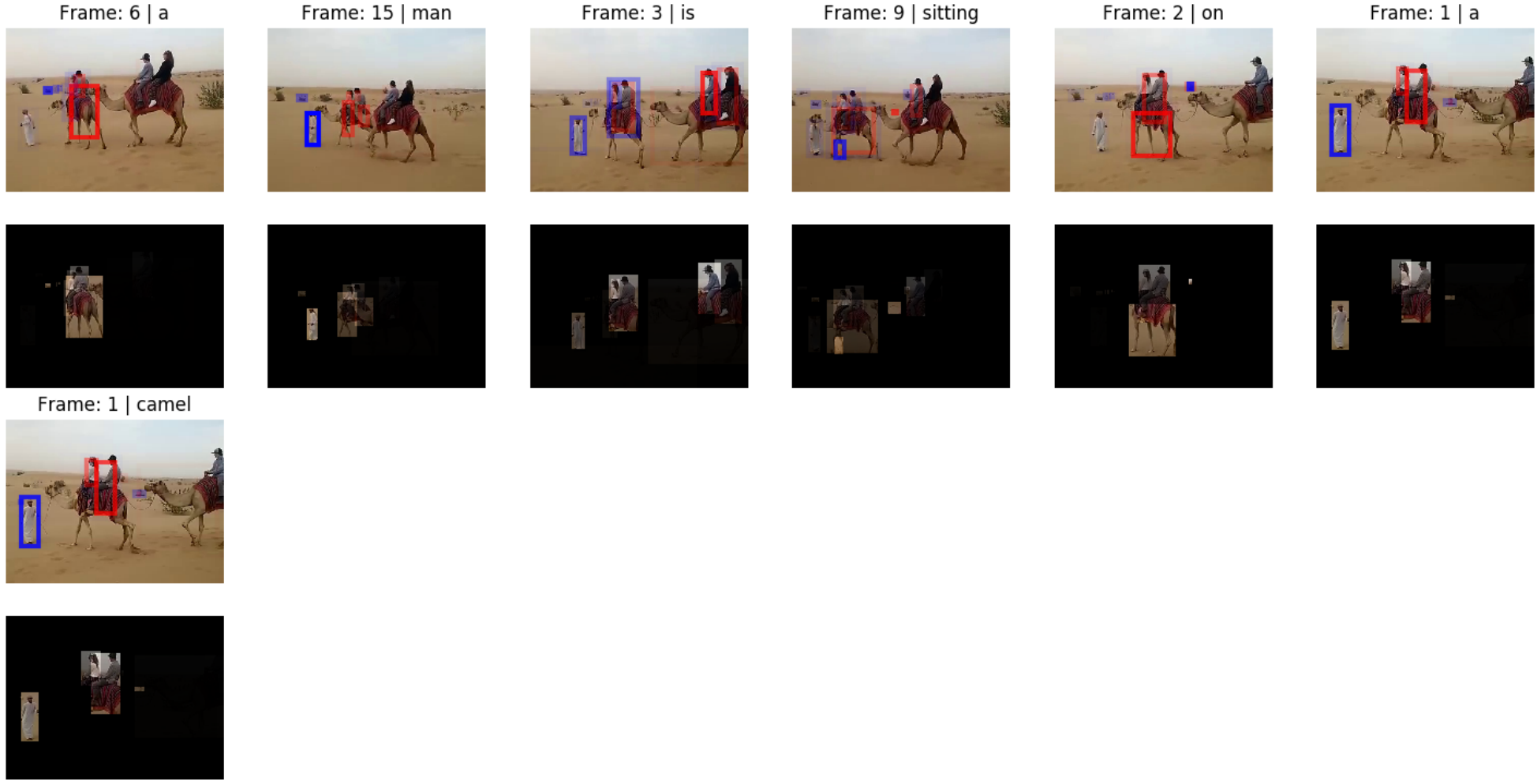}
    \caption{
    \textit{A man is sitting on a camel.}
    The \netshort\ is able to detect the ROIs containing both persons and the camel. 
    We can also observe that it highlights both the ROIs for persons who sit on the camel and the camel itself at frame 3 and 9. 
    However, the proposed method failed to identify that there are multiple people sitting on two camels. Furthermore, in some cases, it selects the person who leads the camels. 
    This seems to be because the same video is also annotated with another caption focusing on that particular person: \textit{A short person that is leading the camels turns around.} 
    }
    \label{fig:camel}
    \vspace{-0.1in}
\end{figure}

To further validate the proposed method, we qualitatively show how the \netshort\ selectively attends to various video frames (temporal), objects, and object relationships (spatial) during each word generation. 
Several examples are shown in Figure~\ref{fig:water_surfing}, \ref{fig:racquetball}, \ref{fig:water_surfing_1}, and \ref{fig:camel}. 
Note that each word generation step, the \netshort\ uses the weighted sum of the video frame representations and the weighted sum of object interactions at corresponding timesteps (co-attention). 
Also, since we aggregate the detected object interactions via LSTM cell through time, the feature representation of the object interactions at each timestep can be seen as a fusion of interactions at the present and past time. 
Thus, if temporal attention has highest weight on $t=3$, it may actually attend to the interaction aggregated from $t=1$ to $t=3$. 
Nonetheless, we only show the video frame with highest temporal attention for convenience. 
We use \textcolor{red}{red} and \textcolor{blue}{blue} to represent the two selected sets of objects ($K=2$). 

In each of the figures, the video frames (with maximum temporal attention) at different timesteps are shown along with each word generation.
All ROIs in the top or bottom images are weighted with their attention weights. In the top image, ROIs with weighted bounding box edges are shown, whereas, in the bottom image, we set the transparent ratio equal to the weight of each ROI. 
The brighter the region is, the more important the ROI is. 
Therefore, less important ROIs (with smaller attention weights) will disappear in the top image and be completely black in the bottom image. 
When generating a word, we traverse the selection of beam search at each timestep.

